\documentclass{article}

\usepackage{microtype}
\usepackage{graphicx}
\usepackage{subfigure}
\usepackage{booktabs} 
\usepackage{multirow}
\usepackage{hyperref}

\usepackage[accepted]{icml2023}

\usepackage{amsmath}
\usepackage{amssymb}
\usepackage{mathtools}
\usepackage{amsthm}

\usepackage[capitalize,noabbrev]{cleveref}

\theoremstyle{plain}

\theoremstyle{definition}

\theoremstyle{remark}

\usepackage[textsize=tiny]{todonotes}

\icmltitlerunning{Instruction-ViT: Multi-Modal Prompts for Instruction Learning in ViT}

\begin{document}

\twocolumn[
\icmltitle{Instruction-ViT: Multi-Modal Prompts for Instruction Learning in ViT}



\icmlsetsymbol{equal}{*}

\begin{icmlauthorlist}
\icmlauthor{Zhenxiang Xiao}{sch1,equal}
\icmlauthor{Yuzhong Chen}{sch1,equal}
\icmlauthor{Lu Zhang}{dept1}
\icmlauthor{Junjie Yao}{sch1}
\icmlauthor{Zihao Wu}{sch2}
\icmlauthor{Xiaowei Yu}{dept1}
\icmlauthor{Yi Pan}{sch4}
\icmlauthor{Lin Zhao}{sch2}
\icmlauthor{Chong Ma}{sch5}
\icmlauthor{Xinyu Liu}{dept2}
\icmlauthor{Wei Liu}{dept4}
\icmlauthor{Xiang Li}{dept3}
\icmlauthor{Yixuan Yuan}{dept2}
\icmlauthor{Dinggang Shen}{sch3,comp1,comp2}
\icmlauthor{Dajiang Zhu}{dept1}
\icmlauthor{Tianming Liu}{sch2}
\icmlauthor{Xi Jiang}{sch1}
\end{icmlauthorlist}

\icmlaffiliation{sch1}{School of Life Science and Technology, University of Electronic Science and Technology of China, Chengdu 611731, China}
\icmlaffiliation{sch2}{School of Computing, The University of Georgia, Athens 30602, USA}
\icmlaffiliation{sch3}{School of Biomedical Engineering, ShanghaiTech University, Shanghai 201210, China}
\icmlaffiliation{sch4}{Glasgow College, University of Electronic Science and Technology of China, Chengdu 611731, China}
\icmlaffiliation{sch5}{School of Automation, Northwestern Polytechnical University, Xi’an 710072, China}

\icmlaffiliation{dept1}{Department of Computer Science and Engineering, The University of Texas at Arlington, Arlington 76019, USA}
\icmlaffiliation{dept2}{Department of Electronic Engineering, Chinese University of Hong Kong, Hong}
\icmlaffiliation{dept4}{Department of Radiotion Oncology, Mayo Clinic, Arizona, USA}
\icmlaffiliation{dept3}{Department of Radiology, Massachusetts General Hospital and Harvard Medical School, Boston 02115, USA}

\icmlaffiliation{comp1}{Shanghai United Imaging Intelligence Co., Ltd., Shanghai 200230, China}
\icmlaffiliation{comp2}{Shanghai Clinical Research and Trial Center, Shanghai, 201210, China}

\icmlcorrespondingauthor{Xi Jiang}{xijiang@uestc.edu.cn}

\icmlkeywords{Machine Learning, ICML}

\vskip 0.3in
]



\printAffiliationsAndNotice{\icmlEqualContribution} 

\begin{abstract}
Prompts have been proven to play a crucial role in large language models, and in recent years, vision models have also been using prompts to improve scalability for multiple downstream tasks. In this paper, we focus on adapting prompt design based on instruction tuning into a visual transformer model for image classification which we called Instruction-ViT. The key idea is to implement multi-modal prompts (text or image prompt) related to category information to guide the fine-tuning of the model. Based on the experiments of several image captionining tasks, the performance and domain adaptability were improved. Our work provided an innovative strategy to fuse multi-modal prompts with better performance and faster adaptability for visual classification models.
\end{abstract}

\section{Introduction}
\label{Introduction}


 It has been a long standing goal of humanity to develop Artificial General Intelligence (AGI) that exhibits human-level or even surpassing intelligence. An essential characteristic of human intelligence is its ability to process information from multiple modalities, which enables individuals to comprehend their surroundings through multiple sources of information and communicate effectively with others \cite{zhao2023brain}. Similarly, artificial intelligence systems are also expected to efficiently handle, integrate, and utilize multimodal data to solve real-world problems. The recent breakthroughs of large language models (LLMs) have provided new insights into realizing this goal. LLMs were initially proposed in the field of Natural Language Processing (NLP) to solve various complex NLP tasks, and have demonstrated remarkable abilities in learning and reasoning. Compared to traditional language models, these LLMs adopt a novel prompt technique which allows the pre-trained LLMs to be adapted to downstream tasks without fine-tuning the models themselves. Through the flexible prompt design, the language model could be pre-trained on massive amounts of raw text and perform few-shot or even zero-shot learning, thus adapting to new scenarios with few or no labeled data \cite{liu2023pre}. For example, the in-context prompt in the GPT series, which is based on auto-regressive pre-training and prompt-based fine-tuning, allowed the model to produce an ideal result for previously unseen tasks without the need to update any parameter \cite{zhang2023makes}.

While large-scale AGI uni-modal (images or texts) models have demonstrated impressive performance in a variety of tasks \cite{brown2020language, kirillov2023segment}, the complexity and diversity of many real-world problems in artificial intelligence often require the integration of information from multiple modalities, such as text, image, and audio. Multi-modal models utilize various methods to fuse data from different modalities, and these methods are often categorized as early (feature) fusion, late (decision) fusion, or intermediate (hybrid) fusion, based on the level in the network at which representations are fused \cite{baltruvsaitis2018multimodal}. However, the choice of fusion method remains highly dependent on the specific domain, data, and task, and there are currently no universal fusion rules. Multi-modal models have shown their potential in enhancing performance across various tasks, such as speaker diarization \cite{gebru2017audio}, text-to-image generation \cite{rombach2022high}, and image description \cite{openai2023gpt4}, but they lack processing methods for some specific tasks like segmentation \cite{kirillov2023segment}.

Recently, several studies have attempted to introduce prompts into visual models so that they can cope with multiple tasks. For example, given a pair of input-output images as the visual prompting for a task example, the model could automatically generate an output image which is consistent with the given examples for a new input image \cite{bar2022visual}, and multiple tasks could be finished as an image inpainting task. The Stable Diffusion \cite{rombach2022high} controls the content of the generated images by inserting text prompts to the latent space. The Segment Anything Model \cite{kirillov2023segment} showed impressive segmentation capabilities on numerous segmentation tasks by adding segmentation prompts which could be points, boxes, text, and masks.

In this work, we concentrate on visual instruction tuning for image captions. The instruction tuning method was first proposed for NLP tasks \cite{wei2021finetuned}. By fine-tuning language models on a collection of datasets described via instructions, the instruction tuning method substantially improves zero-shot performance on unseen tasks. For our design, the instruction tuning method is introduced into the vision transformer (ViT) by adding multi-modal prompts (text or image prompt) related to category information to guide the fine-tuning of the model. Specifically, first, some potential categories of images or text prompts are considered as instruction prompts and contact with the image tokens. Second, the image and instruction prompt tokens are jointly fine-tuned by the model, which we called the Instruction Vision Transformer (Instruction-ViT), to update the representation. Last, a similarity score is calculated between the image CLS token and those potential categories' instruction prompt tokens to finish the image captioning task.

Our contributions and the main findings are summarized as follows:

(1) We introduced instruction tuning to the vision transformer, and both image and text could be used as the instruction prompt to guide tuning.

(2) By conducting experiments on several image captioning tasks, our Instruction-ViT showed better performance and faster adaptability.

\section{Related Work}
\subsection{Large-Scale Multi-Modal Models}
Large-Scale Multi-Modal Models (LMMs) are trained on large-scale datasets to effectively process multiple modalities, particularly for visual-language downstream tasks.
UNITER achieves state-of-the-art performance on various downstream tasks by jointly encoding textual and visual information in a shared representation space \cite{chen2020uniter}.
CLIP utilizes different encoders for images and text, matching them in the latent space to achieve a powerful multi-modal encoder model \cite{radford2021learning}. 
ALIGN also uses a dual-encoder architecture to align visual and language representations by training image-text pairs without manual annotations \cite{jia2021scaling}. 
BLIP pre-trains a multi-modal mixture of encoder-decoder model to tackle both understanding-based and generation-based tasks \cite{li2022blip}.
Flamingo is a family of visual language models trained on large-scale multi-modal web corpora, and can easily adapt to both classification and generation tasks \cite{alayrac2022flamingo}. 
GPT-4, the latest version of GPT models, is a large-scale multi-modal model which is expected to be able to process multiple types of data, including texts, images, audio, and video \cite{openai2023gpt4}. 
Our work also focuses on using pre-trained visual-language models to align language and visual features to better perform visual downstream tasks.


\subsection{Prompt Tuning}
Prompt tuning is a technique used in NLP to improve the performance of language models such as GPT and its other variants \cite{Floridi2020}. Prompt tuning involves fine-tuning a pre-trained language model on a specific task by providing it with a set of prompts or examples relevant to the task \cite{Gu2021ppt}. Unlike conventional fine-tuning, which involves modifying the pre-trained model weights or parameters, prompt tuning requires no changes in the pre-trained model weights. Moreover, prompt tuning becomes increasingly competitive at larger scales, as models with billions of parameters are becoming more common. Existing works show that prompt tuning achieves comparable performance on model tuning, which involves tuning all of the model weights \cite{Liu2022Ptuning, Lester2021}. This is significant because altering the underlying model can be a costly process. In prompt tuning, the model is "frozen" and can be used as is. Furthermore, prompt tuning requires less labeled data compared to other methods.

The goal of prompt tuning is to improve the model's ability to generate high-quality outputs for a specific task, such as text classification or language translation. By training the model on specific prompts or examples, it can learn to generate more accurate and relevant outputs for that task \cite{Liu2023}. Generally, prompt tuning involves providing the model with a set of input-output pairs in the source and target languages, respectively. By fine-tuning the model on these examples, it can learn to generate more accurate translations for new inputs \cite{Liu2022Ptuning}. Prompt tuning is often used in conjunction with other techniques such as transfer learning and data augmentation to further improve the performance of NLP models \cite{Tu2023}.

\subsection{Prompt in Visual Space}
Prompt tuning is proposed to adapt to different downstream tasks, reducing the amount of parameter storage compared to fine-tuning methods while improving the performance on unknown tasks \cite{brown2020language,schick2020exploiting,wei2021finetuned}. Visual Prompt Tuning \cite{jia2022visual} introduces the prompt method into the visual model, which only trains very few parameters to obtain higher classification accuracy than full fine-tuning method. For vision-language models, CoOp adds an additional learnable context prompt to the input of the text encoder to enhance the zero-shot learning capability \cite{zhou2022learning}. To further improve the robustness of the class shift of CoOp, CoCoOp embeds the instance-conditional token obtained on image encoder features from the basis of the context token \cite{zhou2022conditional}. 

\subsection{Instruction Tuning}

Instruction fine-tuning, also known as instruction tuning, is a fine-tuning technique initially introduced for LLMs \cite{wei2021finetuned}. Rather than fine-tuning on a specific downstream task as in BERT-based \cite{devlin2018bert} model tuning, instruction tuning employs data comprising concise instructions and corresponding outputs across a diverse range of tasks and domains. This approach integrates multiple downstream tasks into a single, generic model through a one-time fine-tuning process, considerably reducing total training time and storage space requirements for separate models catering to different tasks. In addition, instruction fine-tuned models not only perform well on the instructions encountered during instruction tuning but also generalize effectively to unseen instructions, thus significantly enhancing zero-shot in-context learning capabilities \cite{ouyang2022training}.

In the field of NLP, based on GPT-3 \cite{brown2020language}, \citeauthor{ouyang2022training} utilizes instruction tuning with RLHF \cite{christiano2017deep} to develop InstructGPT, which better aligns model responses with user intent and minimizes untruthful and toxic content in the output. With further instruction tuning, OpenAI introduces ChatGPT and GPT-4 \cite{openai2023gpt4}, which represent significant strides towards AGI models \cite{zhao2023brain}. Another recently released instruction-tuned model, Alpaca, employs GPT-3.5 to generate a 52k instruction-following dataset \cite{wang2022self} and uses it to fine-tune LLaMA 7b \cite{touvron2023llama}. This achieves comparable performance with GPT-3.5 while requiring a smaller scale and fewer computational resources \cite{alpaca}.

In addition to NLP, recent work has also extended instruction tuning to multi-modal model fine-tuning. \citeauthor{liu2023visual} leverages GPT-4 to generate instruction-following data \cite{peng2023instruction} based on images and the corresponding captions. The resulting LLaVa model demonstrates competitive results with GPT-4 on visual and language understanding tasks.


\section{Methods}
We propose Instruction-ViT, a unified framework to align the input of image and prompts. In this section, we will present the method of creating prompt tokens, provide details of the model's backbone, and discuss the completion of downstream tasks, as well as calculation of loss, respectively. Finally, we introduce our training strategy.

\subsection{Prompt}
We construct the prompt as shown in the bottom right of \cref{MODEL}. In our work, we use the text of class name, the corresponding image, and the combination of text and image as our prompts, respectively. For the text prompt, we use 30 sentence templates in the same way as OpenCLIP \cite{schuhmann2022laion}, of the form like $a\; photo\; of\; a\; \{Class\;Name\}$. After that, we use the pre-trained CLIP text encoder as our prompt encoder, use the 30 constructed prompts as input, and obtain the average result as the prompt token $x_{pt}$. CLIP's pre-trained image encoder is used as the Prompt Encoder for creating image prompt tokens $x_{pi}$. With this encoder, we randomly select an image as the prompt of one class in the training set and finally obtain the prompt tokens. By averaging the text and image prompt tokens together, we obtained the mixed prompt tokens $x_{pm}$ for each class. Therefore, the input prompt tokens can be represented as $x_p=\{x | x \in x_{pt}\;or\;x \in x_{pi}\;or\;x \in x_{pm}\}$ in this work. 

\subsection{Instruction Prompt in Vision Transformer}
\begin{figure}[ht]
\vskip 0.2in
\begin{center}
\centerline{\includegraphics[width=\columnwidth]{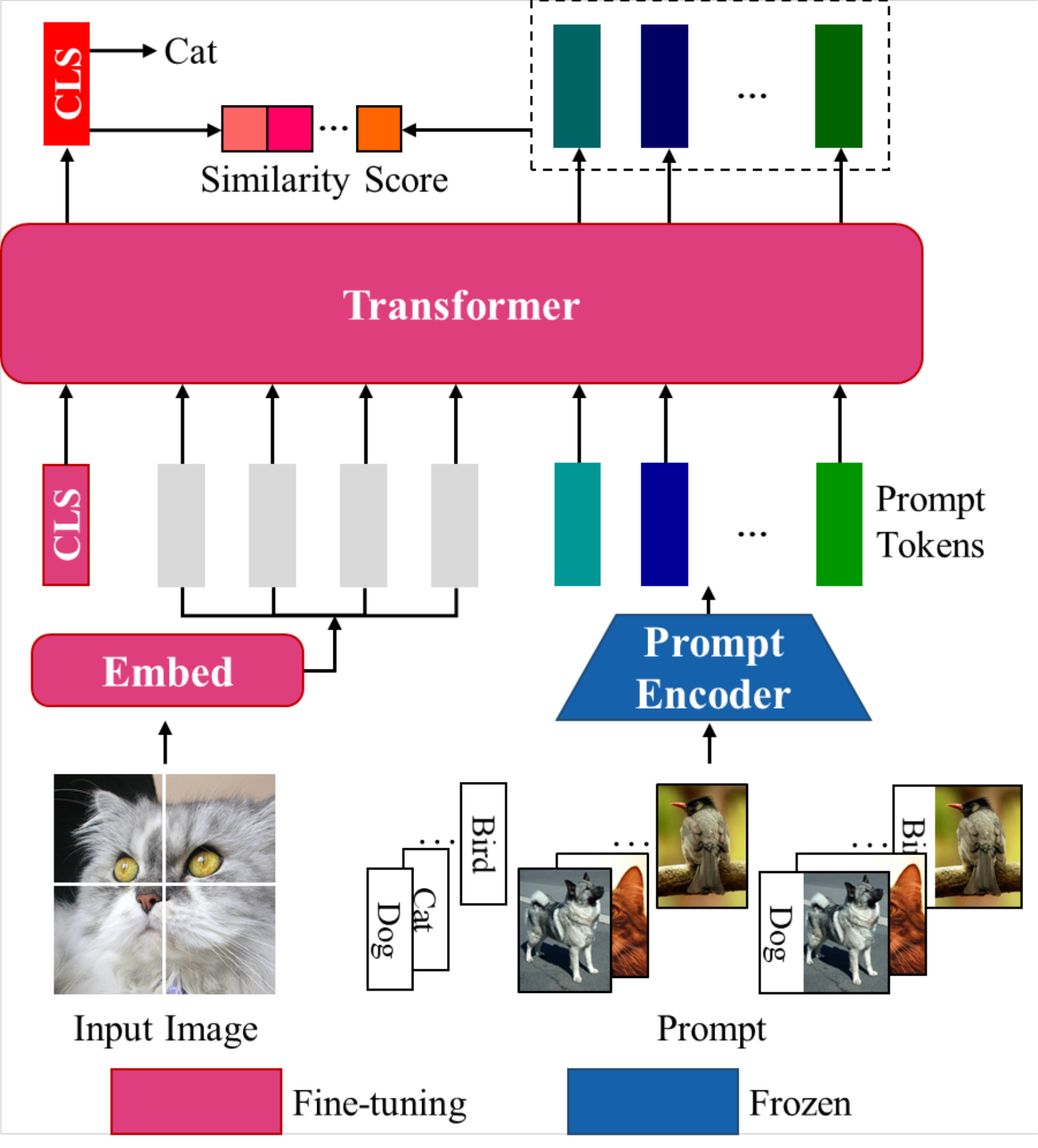}}
\caption{The overall framework of Instruction-ViT. For each image input, the corresponding latent text or visual features are considered as the prompts, by using Transformer's attention mechanism to combine the features of input image and prompts. CLS token is used to complete the downstream task of classification, and the similarity scores computed by CLS and prompt tokens are used to assist in the fine-tuning of the model. At the training stage, the pink module is fine-tuning and the navy blue module keeps frozen.}
\label{MODEL}
\end{center}
\vskip -0.2in
\end{figure}

As shown in \cref{MODEL}, we adopt a ViT as the backbone of our model \cite{dosovitskiy2020image}. For the input of the transformer module, we create a learnable [CLS] token $x_{cls}$ which can represent the global image features and extract prompt features. The other part is the input image, which will be divided into patches and encoded to a sequence of patch embeddings $x_{im}$ by the Embed module. In addition, $x_{cls}$ and $x_{im}$ are added with positional embeddings to retain positional information. The last part is instruction prompt $x_p$, and we can represent the input of our Transformer module as:
$$x_{in}=[x_{cls};x_{im};x_p]$$
where $x_{cls}$, $x_{im}$ and $x_p$ represent the [CLS] token, input image patch embeddings and prompt tokens, respectively. The input $x_{in}$ is then fed into the Transformer module and uses self-attention mechanism of the Transformer module to make [CLS] token utilize features from both $x_{im}$ and $x_p$. In \cref{alg-Instruction-ViT} we show the core implementation of our work.

\subsection{Downstream Task and Loss Construction}
For the final downstream task result, we are able to achieve different downstream task modules. In our work, a classification head is added after the CLS token to accomplish the classification task. For the prediction result $y_{pred}$, we use the cross-entropy loss as the loss function of the classification task, then the loss of classification is:
$$loss_{pred}=CELoss(y_{pred},target)$$
where $target$ is the ground truth, $CELoss$ is the formula to calculate cross-entropy loss.

To improve the alignment of the different modal prompts and the input image, we use the same way as \cite{dosovitskiy2020image} to measure the similarity between the output [CLS] token and prompt tokens by calculating the cosine similarity. The similarity score can be calculated by the formula:
$$Score=y_{cls}y_{p}^T$$
where $y_{cls}$ represents the output of [CLS] token, $y_{text}$ represents the output of prompt tokens, and both $y_{cls}$ and $y_{text}$ are L2 regularized. The class with the largest similarity score with CLS token is the correct target for our expectation, therefore we use the similarity score as the part of loss. The loss can be represented as:
$$loss_{score}=-\log_{}{\frac{\exp{(z_+)}}{{\sum\exp{(z_i)}}}}$$
where $z_+$ represents the similarity score of target sample and $z_i$ represents each similarity score. Then, the final loss can be represented by:
$$loss=loss_{pred}+loss_{score}$$

\subsection{Training Strategy}
\begin{figure}[ht]
\vskip 0.2in
\begin{center}
\centerline{\includegraphics[width=\columnwidth]{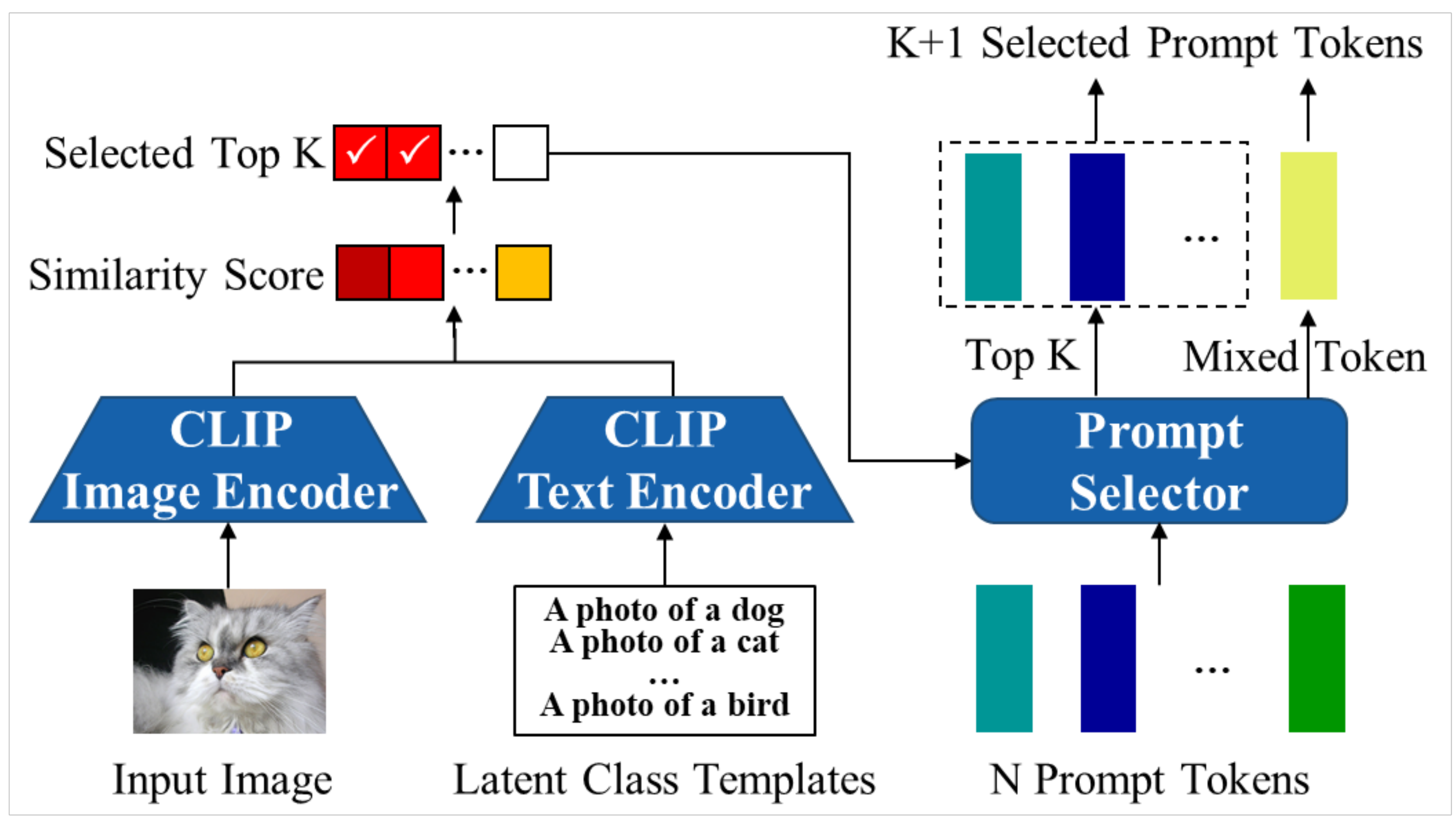}}
\caption{Running mechanism of prompts selected in validation. For an input image of the validation set, feature extraction is performed using the zero-shot CLIP model for the potentially possible class and the image, and its similarity score is calculated. The K prompt tokens with the highest similarity and the average of remaining N-K prompt tokens are selected to next module.}
\label{select}
\end{center}
\vskip -0.2in
\end{figure}

To keep the input image as the main body and minimize the computation time, we limit the number of input prompts in validation. We performed an initial filtering of the potential classes, as shown in \cref{select}. For an input image, we use zero-shot CLIP image encoder and text encoder to extract the feature from input image and text templates representing the latent class information. Similar to the calculation of the final result, we calculate the similarity score by the formula:
$$Score=F_{image}F_{text}^T$$
where $F_{image}$ and $F_{text}$ represents the L2-normalized image and text extracted features. We select K prompts with the highest similarity as the input prompts to the next module. For the other N-K prompt tokens, we compute their average results as the input prompts. In this way, we select K+1 prompt tokens, thus reducing the computation time.

\begin{algorithm}[tb]
   \caption{The PyTorch style code of Instruction-ViT}
   \label{alg-Instruction-ViT}
\begin{algorithmic}
   \STATE class InstructionViT(nn.module):
   \STATE \quad def reset(self, prompt):
   \STATE \qquad self.prompt = self.prompt\_embed(prompt)
   \STATE \quad def forward\_features(self, x):
   \STATE \qquad x = self.to\_patch(x)
   \STATE \qquad x\_cls = self.cls\_token.expand(x.size[0], -1)
   \STATE \qquad x = torch.cat([x\_cls, x], dim=1)
   \STATE \qquad x\_p = self.prompt.expand(x.size[0], -1)
   \STATE \qquad x = torch.cat([x\_p, x], dim=1)
   \STATE \qquad x = self.transformer(x)
   \STATE \qquad return x
   \STATE \quad def forward(self, x):
   \STATE \qquad x = self.forward\_features(x)
   \STATE \qquad pred = self.head(x[:, 0, :])
   \STATE \qquad out\_cls = F.normalize(x[:, 0, :], dim=-1)
   \STATE \qquad out\_p = x[:, -self.prompt.size[0]:, :]
    \STATE \qquad out\_p = F.normalize(out\_p, dim=-1)
   \STATE \qquad Score = torch.einsum('ik, ijk$\rightarrow$ ij', out\_cls, out\_p)
   \STATE \qquad return Score, pred
   
\end{algorithmic}
\end{algorithm}

\section{Experiments}

\subsection{Datasets}

We used 4 well known image classification datasets including Caltech-101 \cite{1384978}, Oxford-III Pets \cite{parkhi2012cats}, Stanford Cars \cite{krause20133d} and Oxford Flowers 102 \cite{4756141}.

\subsection{Implementation Details}
In our work, we adopt the network architecture of 12-layer Transformer blocks with 768 hidden sizes and 12 attention heads in the same way as previous work \cite{dosovitskiy2020image}. For creating prompts tokens, the image and text encoder are adopted from the pre-trained parameters of CLIP. 
In the training stage, the model was trained for 20 epochs (100 epochs in Stanford Cars due to harder classification) with a batch size of 256. We use Adam \cite{kingma2014adam} optimizer with a learning rate 1e-4. We set the foot learning rate to 1e-5, with a linear warm up over the first 5 epochs in the cosine decay strategy. For data augmentation, we adopt RandAugment \cite{cubuk2019randaugment} and Mixup methods \cite{zhang2017mixup}.

\subsection{Evaluation}
We evaluated our model with two methods. The first is fine-tuning the full models in the downstream image classification task. The second is prompt tuning where only the project head and the prompt are learnable while the other parameters are frozen in training. We reported the top-1 accuracy on each dataset task. 

\subsection{Result}

\subsubsection{Fine Tuning Result}
As shown in \cref{table_finetuning}, we compare our model with other models, including ViT \cite{dosovitskiy2020image}, DeiT \cite{touvron2021training}, CaiT \cite{touvron2021going}, PiT \cite{heo2021rethinking}
, ResNet \cite{he2016deep} and EfficientNet \cite{tan2019efficientnet}. As a general observation, the averaged accuracy of our proposed model outperforms other models in fine tuning performance, both ViT-based models, and CNN-based models. The experimental results show that our proposed method can optimize the current ViT-based approach by introducing additional information in prompts.

\subsubsection{Visual Prompt Tuning Result}
We additionally compare the difference of model performance between our training method and the visual prompt tuning (VPT) method in  \cite{jia2022visual}. In this comparison, we keep most of the parameters of the model frozen and fine tune only some of the parameters. In our approach, we fine tune the classification heads and the prompt embedding layer, while the VPT method fine tunes the head and visual prompt proposed. As shown in \cref{table_VPT}, our proposed method is superior to VPT. The results of the experiments demonstrate the feasibility of our proposed methodology for creating special prompts and also prove that the other modal prompt, such as text prompt, can help the completion of visual tasks by our method. In addition, we compare the difference of the accuracy between the three proposed modal prompts. In the four datasets of our experiments, the three modal prompts have their own advantages. In Oxford-III Pets and Oxford Flowers 102, text prompt yields the highest accuracy of 84.74\% and 63.20\%, respectively. In the Caltech-101 dataset, image prompt can achieve the optimal accuracy of 80.85\%, while in the Stanford Cars dataset, the mixed prompt of text and image reaches the optimal accuracy of 66.11\%. These results suggest the importance of using multiple modalities of prompts in different scenarios.

\begin{table*}[t]
\caption{Fine tuning performance in 4 datasets.}
\label{table_finetuning}
\vskip 0.15in
\begin{center}
\begin{small}
\begin{sc}
\begin{tabular}{cccccc}
\toprule
Model & Caltech101 & Pets & Cars & Flowers & Average \\
\midrule
ViT-B                &97.61&94.19&90.90&99.59&95.57   \\
DeiT-B               &96.87&94.71&89.85&96.28&94.43   \\
CaiT-S-24            &96.56&94.27&91.06&96.43&94.58    \\
PiT-B                &96.73&95.29&91.59&97.25&95.22   \\
ResNet-50            &95.55&92.47&87.23&82.93&89.54   \\
ResNet-101           &96.72&93.18&87.60&85.76&90.82   \\
ResNet-152           &97.08&93.04&87.60&88.64&91.59   \\
EfficientNet-b0      &88.74&86.41&83.42&68.98&81.89   \\
EfficientNet-b1      &91.59&87.99&85.10&75.28&84.99   \\
EfficientNet-b2      &93.48&89.17&84.45&78.11&86.30   \\
EfficientNet-b3      &97.59&88.43&84.01&99.59&86.56   \\
Ours(ViT-B)          &97.54&94.19&91.08&99.56&95.59    \\
\bottomrule
\end{tabular}
\end{sc}
\end{small}
\end{center}
\vskip -0.1in
\end{table*}

\begin{table*}[t]
\caption{Fine tuning performance using VPT training strategy in 4 datasets.}
\label{table_VPT}
\vskip 0.15in
\begin{center}
\begin{small}
\begin{sc}
\begin{tabular}{ccccccc}
\toprule
Model &Prompt& Caltech101 & Pets & Cars & Flowers & Average \\
\midrule
ViT-B       &       &73.83&81.69&46.93&58.76&65.30\\
Ours (ViT-B)&text   &79.81&84.74&65.75&63.20&73.38\\
Ours (ViT-B)&image  &80.85&84.58&66.06&60.42&72.98\\
Ours (ViT-B)&mix    &79.78&84.39&66.11&61.10&72.85\\
\bottomrule
\end{tabular}
\end{sc}
\end{small}
\end{center}
\vskip -0.1in
\end{table*}


\section{Conclusion}

In this work, we introduced Instruction-ViT, a simple and effective approach to align the input and prompts across varying modalities. It utilizes the pre-trained parameters from ViT-B as the backbone and CLIP encoders along with a flexible head module for completing downstream tasks like image classification. We show that Instruction-ViT can effectively use uni-modal prompts (e.g., images or texts) as well as multi-modal prompts (e.g., combined image and text features). Experimental results show that Instruction-ViT optimizes the ViT-based model by incorporating prompts in different modalities, and the prompts in different modalities can enhance the effect of the model with fewer parameter training.

In the future, we plan to improve Instruction-ViT from the following perspectives:

(1) We would like to explore the influence of different modules in our Instruction-ViT framework by using more powerful backbones and prompt generators. For example, we will use the Swim Transformer as the backbone and the BERT as the text prompt generator. 

(2) Due to the flexibility of our proposed prompt approach, we will further explore how to design the prompts to achieve better results with our proposed framework. In addition, we will also test different types of prompts, for example using image descriptions as text prompts or using other modality data as prompts such as audio.

(3) Following the represented work like YOLOS \cite{wang2022yolov7} and SAM \cite{kirillov2023segment}, we will further experiment with the performance of other downstream tasks such as target detection and image segmentation.




\nocite{langley00}

\bibliography{example_paper}
\bibliographystyle{icml2023}

\newpage
\appendix
\onecolumn


\end{document}